\documentclass[letterpaper, 10 pt, conference]{ieeeconf}  % Comment this line out if you need a4paper

\IEEEoverridecommandlockouts                              % This command is only needed if 
\usepackage{amsmath} % assumes amsmath package installed
\usepackage{amssymb}  % assumes amsmath package installed
\usepackage{graphicx}
\usepackage{tikz} 
\usepackage{xcolor}
\usepackage{makecell}
\usetikzlibrary{positioning, fit, backgrounds, arrows.meta, calc}

\usepackage{algorithm}
\usepackage{algpseudocode}

\usepackage{url}

\newcommand{\etal}{{\em et~al.}}

\title{\LARGE \bf
On-Demand Robotic Assembly via Differentiable Geometric Part Repair
}

\author{Millicent Schlafly$^{1}$, Fabio Schaub$^{1}$, Diogo Costa Pais$^{1}$, Luca Lelli$^{1}$, Janne Dvorak$^{1}$, Claire Colmont$^{1}$, \\ Sven Marti$^{1}$, and Mark D. Fuge$^{1}$ % <-this % stops a space
\thanks{$^{1}$ETH Zurich, Zurich, Switzerland}%
}

\begin{document}

\maketitle
\thispagestyle{empty}
\pagestyle{empty}

%%%%%%%%%%%%%%%%%%%%%%%%%%%%%%%%%%%%%%%%%%%%%%%%%%%%%%%%%%%%%%%%%%%%%%%%%%%%%%%%
\begin{abstract}

    Transitioning from a digital design to a robotic assembly process currently requires months of expert manual tuning to reconcile part geometries with robotic constraints. This paper presents an end-to-end, autonomous pipeline for the design and physical construction of bespoke wooden assemblies. A generative AI agent  translates user prompts into initial 3D geometries, balancing the visual fidelity of the design with select physical constraints. The assemblability of the design is further improved by a gradient-based repair stage that backpropagates through a graph attention network surrogate to adjust component geometries. In addition to correcting for disjointed and overlapping components, we demonstrate hardware-specific corrections, differentiably optimizing the geometry of components to enable robot screwdriving for $86.7\%$ of $60$ novel natural language inputs, significantly outperforming prior work by a factor of ten. For ten of the structures, we physically demonstrate assemblability with two UR5e robots. This work marks a meaningful step toward on-demand robotic manufacturing, enabling the rapid production of customized, low-volume goods.

\end{abstract}

%%%%%%%%%%%%%%%%%%%%%%%%%%%%%%%%%%%%%%%%%%%%%%%%%%%%%%%%%%%%%%%%%%%%%%%%%%%%%%%%

\section{Introduction}
\label{sec:introduction}
    The economic promise of on-demand robotic manufacturing hinges on its ability to produce highly customized, bespoke goods at speeds comparable to traditional mass production. However, a critical industry-wide bottleneck persists in the transition from an initial digital design to a final assembled product. Currently, reconciling complex part geometries with the rigid physical constraints of robotic cells—such as reachability, collision-avoidance, and end-effector clearances—is a labor-intensive process governed by manual design for assembly principles~\cite{boothroyd_design_1980}. This disconnect necessitates months of expert manual tuning and specialized knowledge to refine a design into a buildable structure~\cite{reinhart2007economic}, rendering the production of low-volume, customized goods prohibitively slow and expensive. Furthermore, the challenge of on demand robot manufacturing is compounded by the heterogeneity of industrial environments; each manufacturing facility utilizes different robotic hardware and software architectures, which impose unique, site-specific constraints on design feasibility.
    
Despite the challenge, the advent of generative AI---and its ability to translate human inputs into 3D structures---has reinvigorated interest in on demand manufacturing and subsequently research in Design for Robotic Assembly~\cite{goldberg_blox-net_2024,kyaw_speech_2025,pun_brickgpt_2025,liu_prompt--product_2025, khendry_idfra_2025, smith_hierarchical_2025, xu_stack_2025, wen_bricksim_2026}. 
    \textbf{In this work, we demonstrate how the geometry of a part can be differentiably optimized to enable robotic assembly.} 
    Our approach builds upon prior methodologies that primarily rely on stochastic perturbations~\cite{goldberg_blox-net_2024} or post-hoc filtering~\cite{pun_brickgpt_2025,liu_prompt--product_2025} to alter part arrangements. The necessity for geometric adaptation arises from the stringent constraints associated with complex robotic fastening mechanisms—for example, screwing—and the use of non-uniform components. Beyond generalizable feasibility criteria such as physical overlaps and structural stability, many designs remain unfeasible due to mechanical bottlenecks like insufficient clearance for the robot gripper and screwdriver or part thicknesses that exceed available screw lengths. Addressing these issues nearly always require fundamental changes to the part geometries themselves rather than only adjustments to their poses.

    Learning-based methods offer a robust framework for adjusting part geometries for assemblability, providing a unified representation to account for diverse assembly failure modes. While computational design optimization is well-established in other domains, its application to the specific constraints of robotic assembly has been limited. Our methodology involves training a surrogate model to map part geometries to assembly success probabilities, subsequently leveraging the model’s gradients to locally optimize designs for enhanced assemblability. Provided a high-fidelity simulation of the manufacturing cell is available for generating labeled training data (e.g., via Gazebo, MuJoCo, Isaac Sim, etc.), our framework is extensible to varied robotic setups. 
    
    To investigate the benefits of geometry optimization for improving robot assembly outcomes, we develop an end-to-end autonomous pipeline for the on-demand robotic assembly of novel wooden objects. In contrast to prior work that assembles objects using pick-and-place operations, our robot manufacturing setup includes a robotic screwdriver and an additional robot arm, significantly complicating robot assembly and consequently imposing additional constraints on the feasible design space. Nevertheless, we show that we can learn critical aspects of assemblability, optimizing over the geometry of the parts to enable robot assembly. We validate our end-to-end autonomous pipeline for generating, editing, and assembling novel objects on ten unseen natural language prompts. Our novel approach to differentiably optimizing part geometries results in the successful modification of $52/60$ unseen inputs (and $22/60$ satisfying all cell constraints end-to-end), significantly outperforming Blox-Net~\cite{goldberg_blox-net_2024}, which was only able to produce $5/60$ designs that passed the target assembly checks (and $4/60$ for end-to-end assembly). By leveraging a differentiable surrogate model to suggest geometric changes to part designs, this work makes significant strides toward the automated production of customized goods.

    % The contributions of this work are:
    % \begin{itemize}
    %     \item an end-to-end autonomous pipeline for generating, editing, and assembling novel objects from wood, [10 examples]
    %     \item a scalable approach for differentiably optimizing the geometries of unseen objects for robot assembly, and [it works for 86.7\% of designs]
    %     \item benchmarking and ablation studies demonstrating that our approach outperforms Blox-Net by differentiably modifying the geometries of the designs for robot screwdriving. 
    % \end{itemize}

%===============================================================================

\section{Related Work}
\label{sec:relatedwork}

\subsubsection{Automating Robot Manufacturing With Fixed Components}
    Recent research has made notable strides in building systems that bridge the gap between natural language prompts and robotic assembly. Projects like Blox-Net~\cite{goldberg_blox-net_2024} and the work of Kyaw \etal~\cite{kyaw_speech_2025} have introduced frameworks where Vision-Language Models generate assembly sequences and instructions for physical construction. Similarly, BrickGPT~\cite{pun_brickgpt_2025} and Prompt-to-Product~\cite{liu_prompt--product_2025} utilize generative AI to create physically stable structures from text. While we achieve a similar goal---turning imaginative ideas into physical products---our approach involves modifying part geometries, which we find to be essential to enabling constuction with more complex assembly mechanisms. 
    
Concurrently, progress is being made in automating multi-step robot assembly planning. AutoMate~\cite{tang_automate_2024} presents a learning framework that uses simulation to train specialist and generalist assembly policies over a diverse dataset of 100 geometries, and Fabrica~\cite{tian_fabrica_2025} tackles the dual-arm assembly of general multi-part objects, expanding to multistep assemblies through the simultaneous optimization of sequences and grasp location. While AutoMate and Fabrica rely on simulation to acquire zero-shot policies for each assembly, BrickCraft~\cite{yu_brickcraft_2026} learns visuomotor skills from 692 demonstrations that are then applied to unseen structures. Additionally, complementary approaches address the inference of sequence steps by modeling it as a graph, shape-matching, or sequence-to-sequence problem, allowing systems to autonomously deduce feasible assembly instructions directly from digital blueprints or visual observations~\cite{atad_efficient_2023, zakka_form2fit_2020, zhu_multi-level_2024}. Our work stands adjacent to the prior papers, improving robot assembly performance primarily through hardware-specific design changes. 

\subsubsection{Design for Robot Assembly Through Human Alterations or Trial-and-Error}

    The design of a product affects the success rates and speed of a manufacturing process~\cite{boothroyd_design_1980}. Until recently~\cite{goldberg_blox-net_2024}, methods for improving assembly outcomes required a human designer with extensive knowledge of the manufacturing process to manually apply documented principles~\cite{chang1997computer,bi2020computer}, meaningfully delaying manufacturing~\cite{reinhart2007economic}. Without an efficient and automated mechanism for changing designs to be assemblable, researchers working in automating robot assembly must limit the scope of which mechanisms are tested according to the specific hardware available~\cite{tian_fabrica_2025}. 
       
    Existing autonomous design for robot assembly methods rely on stochastic perturbation and post-hoc filtering to ensure buildability. The IDfRA framework~\cite{khendry_idfra_2025} and Blox-Net~\cite{goldberg_blox-net_2024} evaluate design stability through iterative trial-and-error rather than proactive geometric modification. Furthermore, most current research restricts designs to modular, discrete units—such as lattices~\cite{smith_hierarchical_2025} or standard blocks~\cite{xu_stack_2025}. These systems often require specialized, hard-coded simulators like BRICKSIM~\cite{wen_bricksim_2026} to handle complex interlocking mechanics, which limits their application to bespoke, non-modular geometries. Our approach differs from prior methods by considering the component geometry as a continuous, optimizable variable.
    
    % Our pipeline departs from these constraints by treating the geometry of bespoke wooden components as a continuous, optimizable variable. By leveraging a surrogate model, we provide the gradients necessary to systematically deform part shapes, ensuring they satisfy specific hardware requirements such as robotic tool and gripper clearance.

\subsubsection{Surrogate Modeling With Graph Neural Networks}
GNNs have proven effective as surrogates for physical and combinatorial systems, learning forward dynamics on particle graphs~\cite{pmlr-v119-sanchez-gonzalez20a} and unstructured meshes~\cite{pfaff2021learning} 1--2 orders of
magnitude faster than the underlying solver, transferring to engineering design~\cite{Wong_2022}, and approximating constraint
satisfaction problems~\cite{10.3389/frai.2020.580607}. We adopt a GNN with attention~\cite{velickovic2018graph} as
our surrogate for part-geometry assemblability.

%===============================================================================

\section{Methods}
\label{sec:methods}

\begin{figure*}[th]
\centering\includegraphics[width=\linewidth]{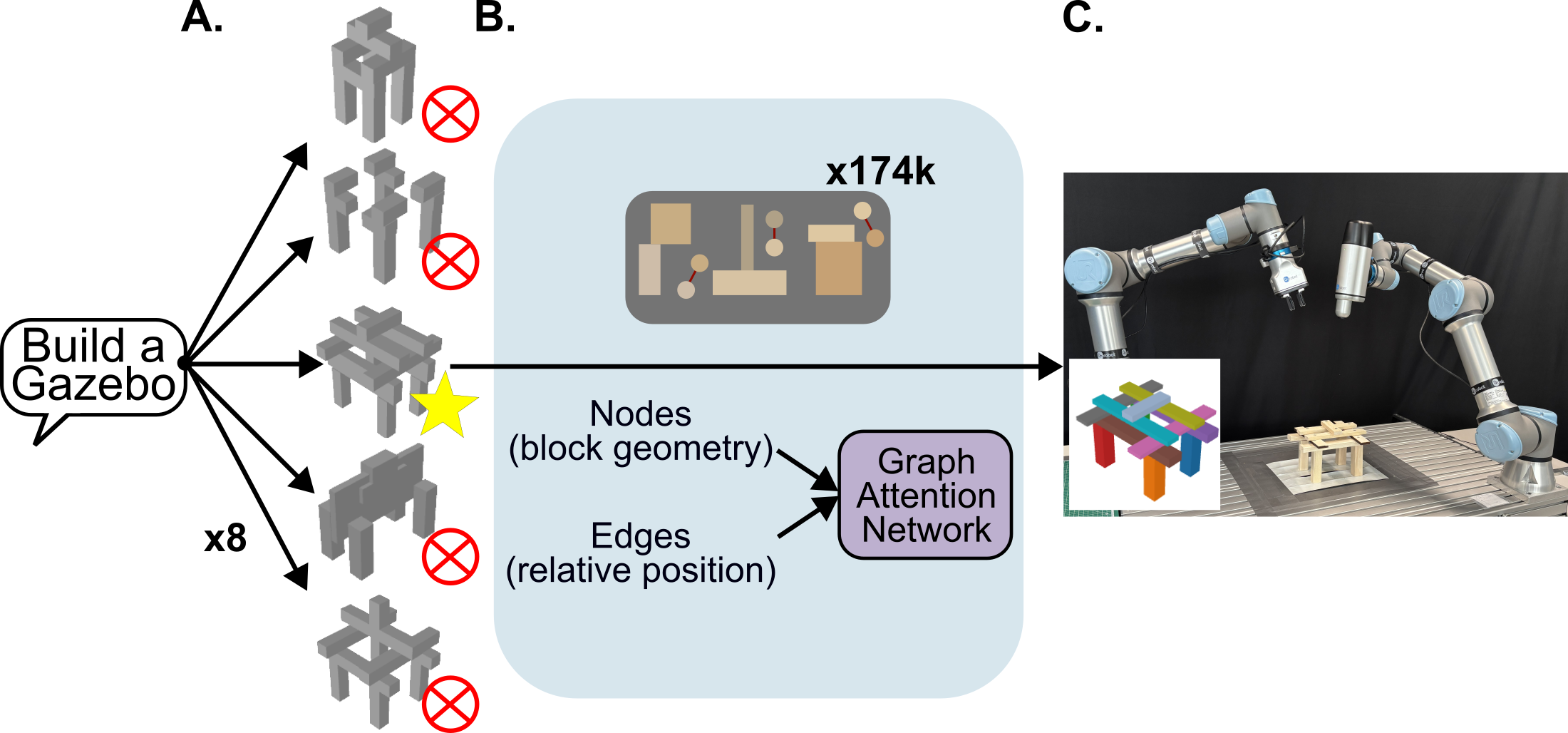}
\caption{\textbf{Woodworking On-Demand Pipeline.} \textbf{A.} A generative AI agent interprets a natural language input to produce $8$ initial geometries, of which many fail due to physics and motion planning checks. \textbf{B.} A differentiable graph attention network, trained from $174,000$ block configurations labeled in NVIDIA Isaac Sim, performs gradient-based optimization to satisfy hardware-specific constraints, such as screwdriver clearance. \textbf{C.} Two UR5e robots physically assemble the optimized design to validate the framework's feasibility for bespoke parts.}
\label{fig:pipeline}
\end{figure*}

Our fully autonomous pipeline in Fig.~\ref{fig:pipeline} for creating and building assemblies from a textual input consists of three steps. Initially, a generative AI agent (e.g., Gemini) produces a comprehensive description of the object and, using the object description, creates $8$ initial designs, selecting from a library of blocks, positioning the blocks and screws, and accounting for some high-level manufacturing considerations such as the order in which parts are assembled (which are further explained in Appendix~\ref{app:system_arch}). Each design is checked for structural stability in the Isaac Sim physics simulator~\cite{liang2018gpu} and motion planning feasibility via MoveIt~\cite{gorner2019moveit}. Of the remaining feasible initial designs, the design with the best prompt satisfaction, as assessed by the AI agent, is selected. Next, we differentiably optimize the design parameters, consisting of the size and position of the blocks, for each assembly operation. The gradients of assemblability probability with respect to the design parameters are obtained from a graph attention network (GAT), which was trained on a wide range of simulated block assemblies. If the design still fails the assemblability checks, the final block positions are perturbed. For the generative AI and block perturbation steps, we replicate the Blox-Net approach presented by~\cite{goldberg_blox-net_2024}, adding a motion planning feasibility check during the initial design generation. We improve upon Blox-Net by differentiably optimizing the part geometries, which proved essential to successful robotic assembly.

In Section~\ref{sec: physical setup}~and~\ref{sec: simulation}, we describe our physical and simulated setup, respectively, explaining the constraints on which designs are assemblable as well as how we obtain labeled assemblablility data. Our approach to modeling and differentiably optimizing the geometries of parts for robot assembly is explained in Section~\ref{sec:gat}~and~Section~\ref{sec:optimization}, respectively. $
\chi^2$ tests in $R$ evaluate statistical significance.% ($\alpha<0.05)$. 

\subsection{Design Constraints Imposed by Physical Setup}
\label{sec: physical setup}

In comparison to the pick-and-place assembly processes evaluated in prior work~\cite{kyaw_speech_2025,goldberg_blox-net_2024,pun_brickgpt_2025}, our robot assembly cell consists of two UR5e robot arms and an OnRobot robot screwdriver, meaningfully constraining which designs are possible to assemble. For each assembly step, the next part is held with a two-finger robot gripper while the robot screwdriver fastens the part to the existing structure. Each part must have a feasible grasp location that is narrower than the $35\,\text{mm}$ grip width of the OnRobot 2FG7 two-finger gripper and does not collide with the table, the existing structure, or the robot screwdriver in its final assembled state. To ensure successful fastening, the thickness of each part where the screw is placed must be less than $20\,\text{mm}$ (since the OnRobot screw feeder only supports screws $25\,\text{mm}$ in length) and the intersection plane of subsequent parts must have dimensions greater than $10\,\text{mm}$ (to ensure the screw is submerged in both parts). Moreover, the robot screwdriver must not collide with either the table or the existing structure. Finally, it must be possible to find a motion plan for each step in the process with our existing software stack. Part detection occurred using NVIDIA's Foundation Pose~\cite{foundationposewen2024,AbbasFoundationPoseROS2} using automatically generated STL files of each object in the assembly. The sequence of high-level motion primitives and initial approximate pose locations of each part is provided by a YAML file generated by the generative AI agent. Grip locations are found deterministically by a script that leverages the known sizes and symmetries of each block.

% \begin{wrapFig.}{r}{0.3\textwidth}
%     \centering
%     \includegraphics[width=\linewidth]{Figures/experiment_setup.png}
%     \caption{\textbf{Experimental Setup.} Common assembly failures arising from the screwing mechanism include insufficient overlap between adjacent pieces, screwdriver collision with the table, and parts that are thicker than the available screw size. }
%     \label{fig:experiment_setup}
% \end{wrapFig.}

\subsection{Dataset Creation and Assessing Assemblability}
\label{sec: simulation}

To obtain labeled data for training, we create a simulated replica of our experimental setup in NVIDIA Isaac Sim. Given a randomized block positions, sizes, and screw locations, we detect whether the assembly succeeded (i.e., whether all of the parts are in the desired locations after assembly). We further detect which objects collided, providing more granular information about how the assembly failed. To enable convenient parallelization over hundreds of environments and faster data collection, evaluation was performed without motion planning or the robot gripper; instead, each block and the robot screwdriver is teleported to the necessary position. Our dataset consists of $2$-block structures over uniformly randomized block sizes in the range $[7,\,200]~\text{mm}$ and block relative positions offset by $[-0.5,0.5]$ (relative to parent block dimensions). Our simulator is further used to evaluate the assemblability of design candidates during both the generative AI and part perturbation steps. 

\subsection{Feasibility Surrogate}
\label{sec:gat}

While rectangular block overlap and component thickness can be evaluated in closed form analytically, we employ a learned GAT surrogate to establish a unified, differentiable framework. This formulation allows gradient backpropagation through complex geometric constraints and is designed to scale to non-orthogonal geometries, intricate joint profiles, and non-differentiable hardware performance metrics where analytical representations become computationally intractable.

% \begin{figure*}[ht]
%     \centering
%     \includegraphics[width=\linewidth]{Figures/Bild.png}
%     \caption{\textbf{}}
%     \label{fig:gat}
% \end{figure*}

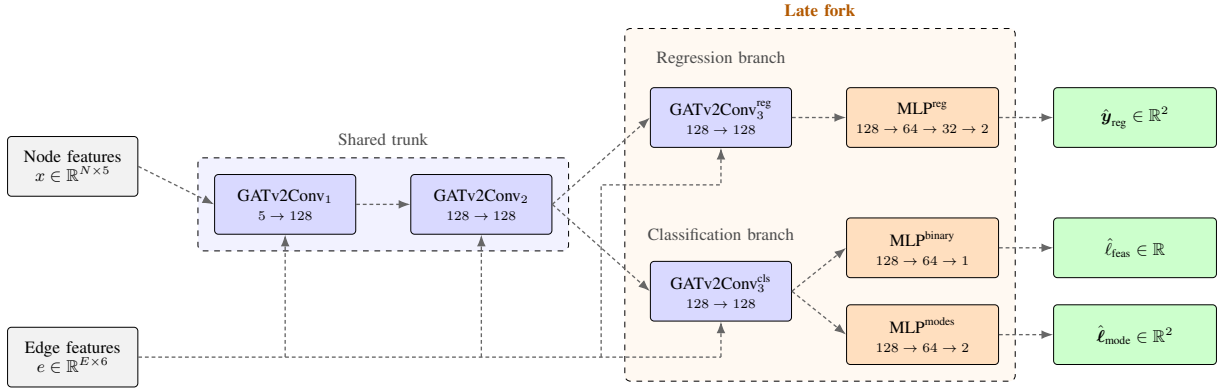
\begin{figure*}[t]
\centering
\resizebox{.9\textwidth}{!}{%
\begin{tikzpicture}[
    font=\small,
    >=Latex,
    node distance=8mm and 10mm,
    feat/.style ={draw, rounded corners=2pt, fill=gray!10,
                  minimum height=11mm, minimum width=24mm, align=center},
    gat/.style  ={draw, rounded corners=2pt, fill=blue!15,
                  minimum height=11mm, minimum width=26mm, align=center},
    mlp/.style  ={draw, rounded corners=2pt, fill=orange!25,
                  minimum height=11mm, minimum width=28mm, align=center},
    output/.style={draw, rounded corners=2pt, fill=green!20,
                  minimum height=11mm, minimum width=30mm, align=center},
    grp/.style  ={draw, dashed, rounded corners=3pt, inner sep=3mm},
    lbl/.style  ={font=\small\color{black!70}},
    flow/.style ={->, dashed, draw=black!60, line width=0.7pt, >=Latex,
                  dash pattern=on 2pt off 1.5pt}
]

% --- Inputs ---
\node[feat] (nx) {Node features\\$x \in \mathbb{R}^{N\times 5}$};
\node[feat, below=24mm of nx] (ex) {Edge features\\$e \in \mathbb{R}^{E\times 6}$};

% --- Shared trunk ---
\node[gat, right=14mm of nx, yshift=-7mm] (g1) {GATv2Conv$_1$\\\scriptsize $5 \to 128$};
\node[gat, right=of g1] (g2) {GATv2Conv$_2$\\\scriptsize $128 \to 128$};

% --- Late fork: regression branch ---
\node[gat, right=18mm of g2, yshift=16mm] (g3r) {GATv2Conv$_3^{\text{reg}}$\\\scriptsize $128 \to 128$};
\node[mlp, right=of g3r] (mr) {MLP$^{\text{reg}}$\\\scriptsize $128 \to 64 \to 32 \to 2$};
\node[output, right=of mr] (or) {$\hat{\boldsymbol{y}}_{\text{reg}} \in \mathbb{R}^2$};

% --- Late fork: classification branch ---
\node[gat, right=18mm of g2, yshift=-16mm] (g3c) {GATv2Conv$_3^{\text{cls}}$\\\scriptsize $128 \to 128$};
\node[mlp, right=of g3c, yshift=8mm]  (mb) {MLP$^{\text{binary}}$\\\scriptsize $128 \to 64 \to 1$};
\node[mlp, right=of g3c, yshift=-8mm] (mm) {MLP$^{\text{modes}}$\\\scriptsize $128 \to 64 \to 2$};
\node[output, right=of mb] (ob) {$\hat{\ell}_{\text{feas}} \in \mathbb{R}$};
\node[output, right=of mm] (om) {$\hat{\boldsymbol{\ell}}_{\text{mode}} \in \mathbb{R}^2$};

% --- Branch labels ---
\node[lbl, above=2.5mm of g3r] (lbl_reg) {Regression branch};
\node[lbl, above=2.5mm of g3c] (lbl_cls) {Classification branch};

% --- Group boxes ---
\begin{scope}[on background layer]
  \node[grp, fill=blue!5,   fit=(g1)(g2)] (trunk) {};
  \node[grp, fill=orange!5, fit=(g3r)(g3c)(mr)(mb)(mm)(lbl_reg)(lbl_cls)] (fork) {};
\end{scope}

\node[lbl, above=1mm of trunk.north] {Shared trunk};
\node[lbl, above=1mm of fork.north, font=\small\bfseries\color{orange!70!black}]
      {Late fork};

% --- Data-flow arrows (all dashed, same color) ---
\draw[flow] (nx.east) -- (g1.west);
\draw[flow] (g1) -- (g2);
\draw[flow] (g2.east) -- (g3r.west);
\draw[flow] (g2.east) -- (g3c.west);
\draw[flow] (g3r) -- (mr);
\draw[flow] (mr)  -- (or);
\draw[flow] (g3c.east) -- (mb.west);
\draw[flow] (g3c.east) -- (mm.west);
\draw[flow] (mb) -- (ob);
\draw[flow] (mm) -- (om);

% --- Edge-feature broadcast (same dashed style) ---
\coordinate (bus_y) at (ex.east);

% Bus line (no arrowhead — taps below carry the heads)
\draw[dashed, draw=black!60, line width=0.7pt, dash pattern=on 2pt off 1.5pt]
    (ex.east) -- (g3c.south |- bus_y);

% Tap-ups into each GAT
\draw[flow] (g1.south  |- bus_y) -- (g1.south);
\draw[flow] (g2.south  |- bus_y) -- (g2.south);
\draw[flow] (g3c.south |- bus_y) -- (g3c.south);

% g3r tap routes around g3c on its left side
\coordinate (riser_x) at ($(g2.east)!0.5!(g3r.west)$);
\coordinate (mid_y)   at ($(lbl_cls.north)!0.5!(g3r.south)$);
\draw[flow] (riser_x |- bus_y) -- (riser_x |- mid_y)
                                -- (g3r.south |- mid_y)
                                -- (g3r.south);

\end{tikzpicture}%
}
\caption{\textbf{Late-fork GAT architecture:} Two shared GATv2Conv
  layers process the assembly graph, then split into a regression branch
  predicting standardized overlap and thickness ($\hat{\boldsymbol{y}}_{\text{reg}}$) and a classification branch
  driving a binary feasibility head ($\hat{\ell}_{\text{feas}}$) and an
  auxiliary failure-mode head ($\hat{\boldsymbol{\ell}}_{\text{mode}}$). Edge
  features are reused as attention inputs at every GAT layer with hidden width
  128, 4 attention heads and dropout 0.1.}
\label{fig:gat-arch}
\end{figure*}

\subsubsection{Graph construction}
  We frame feasibility verification as a per-edge prediction problem on graphs of connected block pairs. %During inference, block connectivity is given explicitly  by the generative AI agent. %We instantiate one graph per declared connected pair and never add edges based on contact or distance. 
  For every connected pair $(b_0, b_1)$, we instantiate a two-node graph with two directed edges. Node features (5-d) are the block's raw axis-aligned edge lengths $[s_x, s_y, s_z]$ concatenated with a one-hot role indicator $[t_0, t_1]$ that marks which node is the anchor
  and which is the candidate. Edge features (6-d) are the center-to-center offset $[\Delta x, \Delta y, \Delta z]$ together with the three axis-pair distances $[d_{xy}, d_{xz}, d_{yz}]$. %All geometric quantities are kept in raw meters. 
  During inference, block connectivity is given explicitly by the generative AI agent.

\subsubsection{Architecture}
Our feasibility surrogate is a graph attention network (GAT)~\cite{velickovic2018graph}, a graph neural network (GNN) variant that uses attention for message passing, as shown in Fig.~\ref{fig:gat-arch}. 
We adopt a late-fork trunk to decouple the regression and classification objectives. Two shared GATv2 layers~\cite{brody2022how} of hidden width $128$ with $4$ attention heads each process the graph. The trunk then splits into a regression branch and a classification branch, each comprising a dedicated third GATv2 layer of the same width.
On top of the regression branch sits a three-layer MLP with GELU activations and hidden widths $64$ and $32$ that predicts $\hat{\boldsymbol{y}}_{\text{reg}} \in \mathbb{R}^2$, the standardized overlap and thickness of the candidate block.
We standardize the regression targets to zero mean and unit variance using statistics from the training split. The overlap target is log-transformed prior to standardization to compress its heavy tail.
Two further two-layer MLPs of hidden width $64$ sit on top of the classification branch: a binary feasibility head producing the logit $\hat{\ell}_{\text{feas}}$, and a multi-label mode head producing $\hat{\boldsymbol{\ell}}_{\text{mode}} \in \mathbb{R}^2$ whose two components independently flag overlap and thickness violations. Read-out happens only at the active node, the anchor node serves purely as context. The full model has $131{,}813$ parameters. Dropout of $0.1$ is applied between layers.

\subsubsection{Training objective}
We minimize a multi-task loss~\cite{Caruana1997MultitaskL} combining a regression term on the two
geometric outputs with binary cross-entropies on the two classification heads,
\begin{equation}
  \mathcal{L}_{\text{tot}}
  =
  \mathcal{L}_{\text{reg}}
  \,+\, \alpha\,\mathcal{L}_{\text{feas}}
  \,+\, \beta\,\mathcal{L}_{\text{mode}},
  \qquad \alpha = \beta = 0.1
  \label{eq:total_loss}
\end{equation}
The regression term is a sum of boundary-weighted~\cite{Lin_2017_ICCV} Huber losses~\cite{Huber1964RobustEO} ($\delta = 1$) over the two standardized targets, with boundary weight near the decision threshold.
The two classification terms are weighted binary cross-entropies~\cite{BUDA2018249} on the
feasibility logit and the two-component mode logit. Positive-class weights are set from the training set frequencies~\cite{BUDA2018249}. The per-mode weights are additionally clipped at $50$ to prevent rare modes from dominating the gradient. The mode targets carry label smooting~\cite{Szegedy_2016_CVPR} with $\varepsilon_{ls} = 0.05$. The mode head is auxiliary\textemdash it shares the classification trunk with the feasibility head during training and is discarded at inference. For further details see Appendix~\ref{app:gat_loss}.

\subsubsection{Training}
We train on $174{,}000$ graphs with a dataset split $80/10/10$ and optimize with Adam~\cite{DBLP:journals/corr/KingmaB14} at learning rate $2 \times 10^{-3}$ and batch size $2048$. The learning rate is decayed by a factor of $0.7$ whenever the validation composite mean absolute error $\tfrac{1}{2}(\mathrm{MAE}_o + \mathrm{MAE}_t)$ plateaus for $12$ epochs (ReduceLROnPlateau, $\eta_{\min} = 10^{-6}$). Training runs for up to $350$ epochs with early stopping (patience $60$) on the same composite metric. The best checkpoint is retained. %For further details see Table~\ref{tab:gat-hparams}.

\subsubsection{Performance}
Our best run converged at epoch~$196$ to a test set feasibility-classification $F_1$ of $0.965$, threshold-derived $F_1$ of
$0.924$ (precision $0.927$, recall $0.922$), and mean absolute errors of $4.4\,\text{mm}$ on overlap and $1.7\,\text{mm}$ on thickness, against thresholds of $10$ and $20$\,mm respectively. The standardized-output standard-deviation ratio between predictions and targets remained close to unity ($0.89$ for overlap, $0.93$ for thickness), confirming that the regression heads have not collapsed onto the mean.

\subsection{Gradient-based assembly repair}
\label{sec:optimization}
To convert infeasible layouts created by the generative AI agent into manufacturable designs, we propose a four-stage repair pipeline. First, two analytical passes resolve pairwise interpenetrations and snap each child block onto its parent's contact face, bringing the assembly into a regime where the surrogate's predictions are reliable.
Second, blocks resting on the table are frozen as anchors, and the remaining blocks are repaired bottom-up via a breadth-first traversal: each child's size $\mathbf{s} \in \mathbb{R}^3$ and position $\mathbf{p} \in \mathbb{R}^3$ are optimized against the already committed parent by Adam on a composite loss combining constraint hinges on the GAT outputs with quadratic drift regularizer. Hard projections after each step bound position drift and per-axis size changes to a trust region in which the surrogate generalizes well, and periodic analytical snaps prevent drift along the contact normal. Finally, fasteners are placed inside each repaired contact rectangle subject to a safety inset and global separation constraints. The full loss formulation and pseudocode are given in Appendix~\ref{app:repair_details}.

\begin{figure*}[th]
    \centering\includegraphics[width=.94\linewidth]{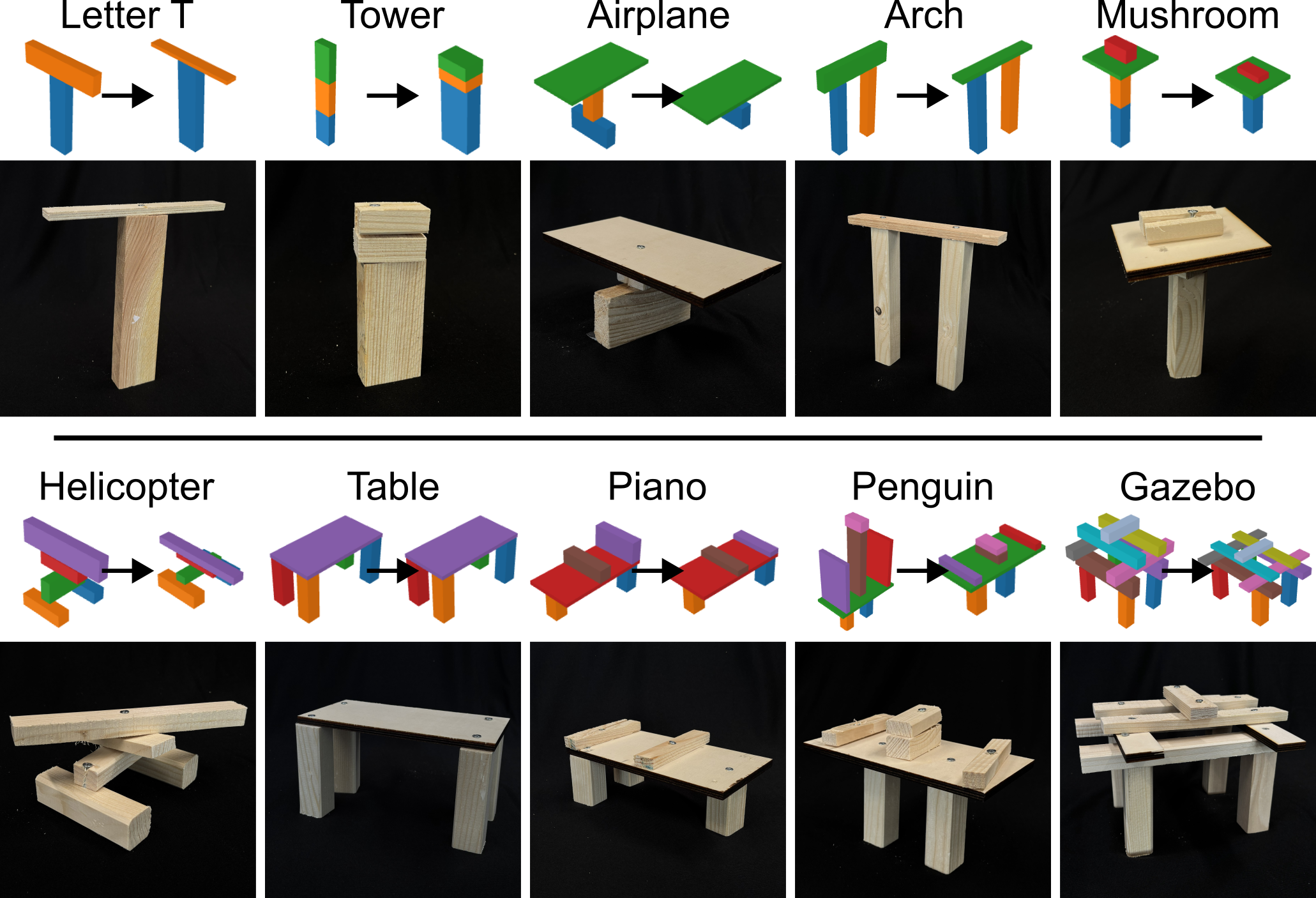}
    \caption{\textbf{Final Assemblies for Ten Natural Language Inputs}. For each prompt, we show the best design candidate produced by the generative AI agent (left) followed by the modified design (right). We physically construct each object to validate assemblability.}\label{fig:results}
\end{figure*}

\section{Experimental Results}
\label{sec:result}

\begin{table*}[ht]
\centering
\small
\begin{tabular}{l|cccc|c|c}
\multicolumn{1}{c}{} % This removes the vertical line after the empty corner cell
 & \multicolumn{1}{c}{\makecell{Gap/\\Concurrence}} 
 & \multicolumn{1}{c}{\makecell{Structural\\Tipping}} 
 & \multicolumn{1}{c}{\makecell{Sufficient\\Overlap}} 
 & \multicolumn{1}{c}{\makecell{Thickness\\}}
 & \multicolumn{1}{c}{\makecell{Overall\\Screwable}}
 & \multicolumn{1}{c}{\makecell{End-to-End\\ (Full System)}}  \\ \hline
Ours & $93.3\%\;(56)$  & $96.7\%\;(58)$ & $\mathbf{98.3\%}\;(59)$ & $\mathbf{86.7\%}\;(52)$ & $\mathbf{86.7\%}\;(52)$ & $\mathbf{36.7\%}\;(22)$ \\
Blox-Net & $\mathbf{100\%}\;(60)$ & $\mathbf{100\%}\;(60)$ & $85.0\%\;(51)$ & $8.3\%\;(5)$ & $8.3\%\;(5)$ & $6.7\%\;(4)$  \\
& $p=0.127$ & $p=0.476$ & \textbf{*}$\mathbf{p<0.05}$ & \textbf{***}$\mathbf{p<0.001}$ & \textbf{***}$\mathbf{p<0.001}$ &  \textbf{***}$\mathbf{p<0.001}$\\ \hline
\end{tabular}
\caption{\textbf{Benchmarking Results From Simulated Assembly Checks.} We present the percentage of $60$ designs that passed each assembly check followed by the number of designs in parentheses. 
}
\label{table:benchmark}
\end{table*}

We validate our end-to-end fully autonomous pipeline for generating, editing, and assembling novel objects with ten natural language inputs in Fig.~\ref{fig:results}. For all ten inputs, our approach modifies the geometry of the blocks if necessary to enable robot assembly, typically shrinking the vertical thickness so that the $25\,\text{mm}$ screw can pass through both adjacent blocks. In some cases, the geometric changes impacted the prompt fidelity of the design, such as for the \textit{penguin} and the \textit{tower}. The \textit{table} required no geometric modifications. Wooden parts were manually cut to size following the gradient-based geometric resizing. During physical construction, wood splitting and inconsistencies in the robot screwdriving mechanism resulted in modest deviation from target assemblies. Across the 10 physical builds, our system achieved 10/10 full simulated feasibility, 7/10 intervention-free full builds, 47/50 intervention-free individual block placement steps, and 10/10 total completions with minor human interventions (three instances of screw feeder jams and three block resets due to cell vibration). For convenience, we manually provided initial location estimates for each component. %Overall, nearly all object remain visually recognizable following physical construction. 

\begin{table*}[ht]
\centering
\small
\begin{tabular}{l|cccc|c}
% \multicolumn{1}{c}{ } & \multicolumn{1}{c}{gap/concurrence} & \multicolumn{1}{c}{structural tipping} & \multicolumn{1}{c}{sufficient overlap} & \multicolumn{1}{c}{screwable} & \multicolumn{1}{c}{overall}  \\ \hline
\multicolumn{1}{c}{} % This removes the vertical line after the empty corner cell
 & \multicolumn{1}{c}{\makecell{Gap/\\Concurrence}} 
 & \multicolumn{1}{c}{\makecell{Structural\\Tipping}} 
 & \multicolumn{1}{c}{\makecell{Sufficient\\Overlap}} 
 & \multicolumn{1}{c}{\makecell{Thickness}} 
 & \multicolumn{1}{c}{\makecell{Overall\\Screwable}}  \\ \hline
Ours & $93.3\%\;(56)$  & $96.7\%\;(58)$ & $\mathbf{98.3\%}\;(59)$ & $\mathbf{86.7\%}\;(52)$ & $\mathbf{86.7\%}\;(52)$ \\
MLP+KD-Tree Cost & $21.6\%\;(13)$ & $71.7\%\;(43)$ & $63.3\%\;(38)$ & $36.7\%\;(22)$ & $11.7\%\; (7)$\\
Ours w/o GenAI & $95.0\%\;(57)$ & $86.7\%\;(52)$ & $93.3\%\;(56)$ & $81.7\%\;(49)$ & $66.7\%\;(40)$\\
Ours w/o Perturbations & $93.3\%\;(56)$ & $96.7\%\;(58)$ & $96.7\%\;(58)$ & $\mathbf{86.7\%}\;(52)$ & $\mathbf{86.7\%}\;(52)$ \\
Ours w/o GAT (Blox-Net) & $\mathbf{100\%}\;(60)$ & $\mathbf{100\%}\;(60)$ & $85.0\%\;(51)$ & $8.3\%\;(5)$ & $8.3\%\;(5)$\\ \hline
\end{tabular}
\caption{\textbf{Ablation Results From Simulated Assembly Checks.} We present the percentage of $60$ designs that passed each assembly check followed by the number of designs in parentheses. For the condition without GenAI, only one candidate design is created (as opposed to $8$). 
% Assembly checks are gap/concurrence, structural tipping, screwable (the part is thin enough for the screw with a fixed length to reach the adjacent structure), sufficient overlap (adjacent parts share enough of the bounded connecting plane for screwing), and screwdriver collision (screwdriver does not collide with table).
}
\label{table:ablation}
\end{table*}

By differentiably optimizing the geometry of the parts, our approach results in significantly fewer screwable and sufficient overlap failures than Blox-Net~\cite{goldberg_blox-net_2024} (screwable: $\chi^2(1,N=60)=70.7, p<2.2e-16$; overlap: $\chi^2(1,N=60)=5.35, p=0.021$), as shown in Table~\ref{table:benchmark}. We evaluated our approach in comparison to Blox-Net in simulation for $60$ natural language prompts: 
\textit{chair}, \textit{human}, \textit{dog}, \textit{boat}, \textit{rabbit},  \textit{shark}, \textit{eiffeltower}, \textit{umbrella}, \textit{car}, \textit{rook chess piece},
\textit{bridge}, \textit{house}, \textit{matterhorn}, \textit{camel}, \textit{bicycle}, \textit{cactus}, \textit{castle}, \textit{sofa}, \textit{taj mahal}, \textit{windmill},
\textit{pyramid}, \textit{lighthouse}, \textit{anchor}, \textit{guitar}, \textit{drum}, \textit{sailboat}, \textit{cake}, \textit{robot}, \textit{snowman}, \textit{cheese},
\textit{pineapple}, \textit{octopus}, \textit{pumpkin}, \textit{bell}, \textit{barn}, \textit{bench}, \textit{bookshelf}, \textit{catapult}, \textit{lantern}, \textit{mailbox},
\textit{pagoda}, \textit{rocket}, \textit{silo}, \textit{stairs}, \textit{teapot}, \textit{telescope}, \textit{wagon}, \textit{wheelbarrow}, \textit{wishing well}, \textit{fence},
and the prompts shown in Fig.~\ref{fig:results}. Blox-Net is successfully able to prevent gaps between adjacent components, overlapping components, and structural tipping more frequently than our approach, although insignificantly (gap/concurrence: $\chi^2(1,N=60)=2.33, p=0.127$; tipping: $\chi^2(1,N=60)=0.51, p=0.476$). However, it is unable to address hardware-specific failure modes such as insufficient overlap between the block faces of adjacent components or insufficient thickness for screwing, both of which proved critical to successful assembly. In contrast, our approach can prevent failures associated with screwing with a $86.7\%$ success rate, significantly higher than Blox-Net ($\chi^2(1,N=60)=70.7, p<0.001$). Likewise, full system end-to-end assembly success, which includes failures due to unmodeled gripper access and motion planning constraints, is significantly higher for our approach than Blox-Net ($\chi^2(1,N=60)=14.19, p<0.001$).

% Code for stats in R 

% # Create a matrix of your success and failure counts
% # Row 1: Group 1 (e.g., 52 successes, 8 failures)
% # Row 2: Group 2 (e.g., 5 successes, 55 failures)
% data_matrix <- matrix(c(52, 8, 
%                         5, 55), 
%                       nrow = 2, byrow = TRUE)
% # Optional: Add names to make the output easier to read
% rownames(data_matrix) <- c("Group_1", "Group_2")
% colnames(data_matrix) <- c("Success", "Failure")

% # View your table
% print(data_matrix)

% # Run the test
% chi_test_result <- chisq.test(data_matrix)

% # Print the results
% print(chi_test_result)

To understand which step in our design process is most impactful, we conducted an ablation study in simulation across the same $60$ natural language inputs in Fig.~\ref{table:ablation}. When we remove the generative AI parallelization, producing only one candidate design instead of eight, assembly success meaningfully declines, particularly for the structural tipping assembly check. Removing the perturbation step has no impact on assembly failures; since screwing involves physical fastening, our physical setup may have been better able to withstand positional deviations during assembly than for the pick-and-place assembly processes from the Blox-Net paper, explaining the absence of benefit. Our method without the differentiable geometric modification step and motion planning check by the generative AI agent is identical to Blox-Net. We also compare to using a MLP surrogate model with an optimization cost on moving away from assemblable designs via a KD-Tree; see Appendix~\ref{app:mlp_baseline} for more details.  
Overall, the geometric optimization step is key to assembly performance. Since we used a pretrained surrogate model, the optimization step is also the fastest part of our design pipeline, taking only around $90\,\text{s}$. For context, on a laptop (Ultra 7 / 16GB RAM / RTX 500 Ada (4GB)), the entire pipeline took an average of $54\,\text{min}$. 

\begin{figure*}[ht]
    \centering\includegraphics[width=.94\linewidth]{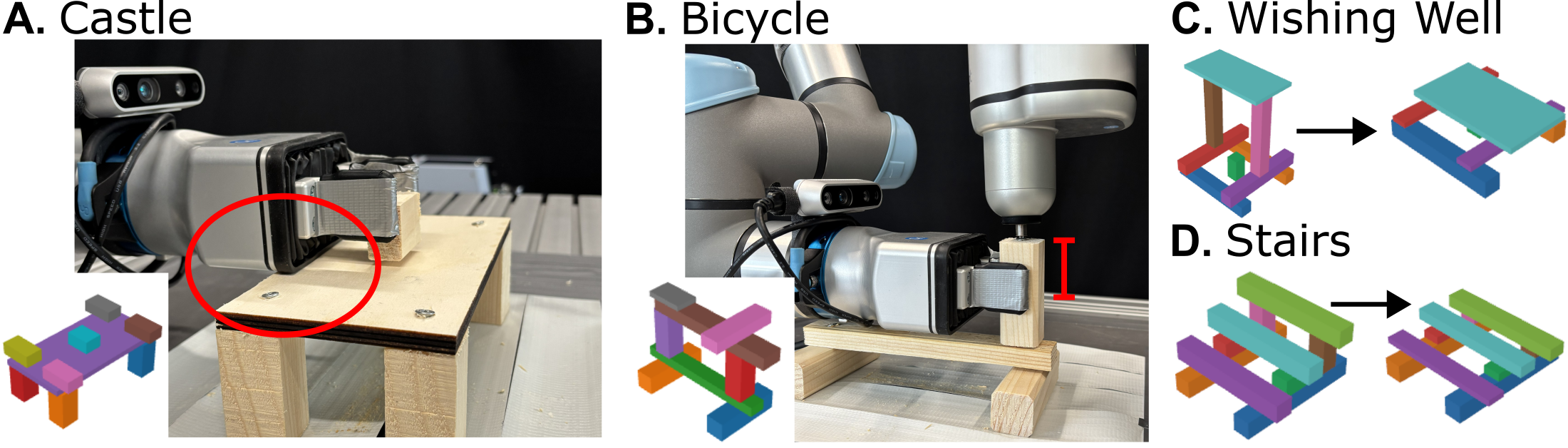}
    \caption{\textbf{Remaining Assembly Failures. } \textbf{A.} For the \textit{castle}, the body of the gripper collides with the remaining assembly. \textbf{B. } The \textit{bicycle} handlebar post remains longer than the available screws after optimization. \textbf{C.} After assemblability modifications for the \textit{wishing well} and \textit{stairs}, the designs are no longer visually recognizable.   
}\label{fig:failures}
\end{figure*}

\section{Limitations}
\label{sec:conclusion}

% This approach has two primary limitations. Firstly, there are a range of assembly failures that are not explicitly considered by our feasibility model such as gripper-screwdriver collision, gripper collision with either the table or the remaining structure, or motion planning feasibility. Consequently, we are not able to sufficiently modify designs like the \textit{castle} in Fig.~\ref{fig:failures}\textbf{A} for assemblability. The central castle tower in unable to be placed due to gripper collision with other parts of the structure. These types of failures were frequent; although $86.7\%$ of designs passes the assembly checks we explicitly optimized for, \textcolor{red}{$63.3\%$ of the $60$ evaluated inputs (leaving $36.7\%$ end-to-end success)} were ultimately not assemblable for other reasons, typically involving the gripper. 
% Although our approach could in principle extend to a wide range of assembly criteria, as the number of design constraints increases, modeling the assemblable subspace with high classification accuracy may become more challenging.  

This approach has two primary limitations. First, the GAT surrogate specifically optimizes for part-level geometric constraints (e.g., screwdriver clearance, component thickness, face overlap), while leaving multi-body kinematic constraints—such as gripper collision with cell components, swept-volume clearance, and global reachability—to post-hoc simulation checks. These unmodeled kinematic constraints, particularly robot gripper collisions during insertion, account for the majority of the gap between our target geometric constraint success rate ($86.7\%$) and the full end-to-end cell build success rate ($36.7\%$). Consequently, we are not able to sufficiently modify designs like the \textit{castle} in Fig.~\ref{fig:failures}\textbf{A} for assemblability, where the central castle tower cannot be placed due to gripper collisions with surrounding components. Although our approach could in principle extend to a wider range of assembly criteria, as the number of design constraints increases, modeling the assemblable subspace with high classification accuracy may become more challenging.

Secondly, since our optimization process is local, our approach is only reasonable if the initial design is close enough to an assemblable design. We found that it was difficult to tune the optimizer to both penalize large design changes and also find solutions in situations where large design changes were necessary. Our approach failed to decease the height of the bicycle in Fig.~\ref{fig:failures}\textbf{B}, rendering the final design unassemblable. Likewise, our approach cannot find solutions like adding/removing parts, reordering assembly sequences, or multi-block single-screw fastening. Due to local, continuous, single factor optimization, some designs are no longer recognizable after modifying for assemblability (e.g., the \textit{wishing well} and \textit{stairs} in Fig.~\ref{fig:failures}\textbf{C}), although one could envision building upon this approach to simultaneously optimize for assemblability along with prompt visual fidelity and functionality metrics.

%===============================================================================

\section{Conclusion}
\label{sec:conclusion_sec}

Designing an assembly for robot manufacturing is a laborious process,  involving the identification of a wide range of possible failure modes that span hardware and software considerations. Instead of having a human designer manually specify the relationship between geometry and robot feasibility, we propose learning the relationship from labeled simulation data. As we demonstrate in this paper, a learned feasibility model is differentiable---enabling fast geometric modifications---and generalizable to dozens of unseen structures. Since our approach can incorporate hardware-specific assemblability criteria, such as those arising from robot screwdriving, it outperforms Blox-Net. Moreover, this approach has the potential to scale up to more particular assemblability requirements like motion planning feasibility without becoming computationally unmanageable or simultaneously optimize for a differentiable stress or compliance objective. Although we use our feasibility model to modify designs produced by generative AI, one could envision using a similar approach to modify novel components for an aircraft or appliance. By presenting an end-to-end fully autonomous design generation, modification and assembly---enabled by differential geometric design optimization---we pave the way for on-demand robot manufacturing with of other manufacturing cells with more complex parts and assemblability criteria.

%===============================================================================

% \clearpage
% The acknowledgments are automatically included only in the final and preprint versions of the paper.
% \acknowledgments{If a paper is accepted, the final camera-ready version will (and probably should) include acknowledgments. All acknowledgments go at the end of the paper, including thanks to reviewers who gave useful comments, to colleagues who contributed to the ideas, and to funding agencies and corporate sponsors that provided financial support.}

%===============================================================================

% \bibliography{references}  % .bib

% \appendix

%===============================================================================

%%%%%%%%%%%%%%%%%%%%%%%%%%%%%%%%%%%%%%%%%%%%%%%%%%%%%%%%%%%%%%%%%%%%%%%%%%%%%%%%



% \section*{ACKNOWLEDGMENT}

% The preferred spelling of the word ÒacknowledgmentÓ in America is without an ÒeÓ after the ÒgÓ. Avoid the stilted expression, ÒOne of us (R. B. G.) thanks . . .Ó  Instead, try ÒR. B. G. thanksÓ. Put sponsor acknowledgments in the unnumbered footnote on the first page.

% %%%%%%%%%%%%%%%%%%%%%%%%%%%%%%%%%%%%%%%%%%%%%%%%%%%%%%%%%%%%%%%%%%%%%%%%%%%%%%%%

% References are important to the reader; therefore, each citation must be complete and correct. If at all possible, references should be commonly available publications.

\bibliographystyle{ieeetr}
\bibliography{references}

\appendices

\section{System Architecture and Motion Planning}
\label{app:system_arch}
The system execution context relies on ROS~2 Jazzy and MoveIt~2 for motion planning. Parts and tasks are generated via structured YAML files output by the generative AI agent. Assembly operations are structured around three primitive steps: \texttt{PICKANDPLACE}, \texttt{PICKANDHOLD}, and \texttt{SCREWPARTS}. Motion trajectories are computed using OMPL's \texttt{RRTConnect} for free-space motion and collision-checked Cartesian paths for alignment tasks. Object detection and pose tracking are performed using FoundationPose initialized via SAM2 segmentations from an Intel RealSense D435 camera. Generative AI designs are guided by a 15-rule system prompt enforcing bottom-to-top assembly order, stable support constraints, gripper clearance limits, and screw allocation per joint.

\section{GAT Surrogate Training Objective}
\label{app:gat_loss}
The GAT surrogate loss $\mathcal{L}_{\text{tot}} = \mathcal{L}_{\text{reg}} + \alpha\mathcal{L}_{\text{feas}} + \beta\mathcal{L}_{\text{mode}}$ ($\alpha=\beta=0.1$) unifies regression and classification goals. Overlap $y_o$ and thickness $y_t$ are standardized as:
\begin{equation}
  \tilde{y}_o^{(i)} = \frac{\log(y_o^{(i)} + \varepsilon_{\log}) - \mu_o}{\sigma_o}, \quad \tilde{y}_t^{(i)} = \frac{y_t^{(i)} - \mu_t}{\sigma_t}
\end{equation}
Regression uses a boundary-weighted Huber loss ($\delta = 1$):
\begin{equation}
  w_k^{(i)} = 1 + g \cdot \exp\!\left( -\,\frac{\bigl|\tilde{y}_k^{(i)} - \tilde{\tau}_k\bigr|}{\sigma_b} \right)
\end{equation}
where $g = 3.0$ and $\sigma_b = 0.5$, heavily weighting samples near the feasibility boundary. Classification heads utilize positive-class weighted binary cross-entropies ($w_{\max}=50$) with target label smoothing ($\varepsilon_{ls} = 0.05$).

\section{Gradient-Based Assembly Repair}
\label{app:repair_details}
Given initial parameters $(\mathbf{s}_0, \mathbf{p}_0)$, candidate geometries are adjusted bottom-up via Adam on the composite loss:
\begin{equation}
    \mathcal{L}_{\text{repair}}(\mathbf{s}, \mathbf{p}) = \mathcal{L}_{\text{feas}}^{\text{hinge}} + \mathcal{L}_{\text{reg}}^{\text{hinge}} + \lambda_s \|\mathbf{s} - \mathbf{s}_0\|_2^2 + \lambda_p \|\mathbf{p} - \mathbf{p}_0\|_2^2
\end{equation}
with drift penalties $\lambda_s = 5, \lambda_p = 1$, and hinge losses scaling through $K = 10^6$:
\begin{equation}
  \mathcal{L}_{\text{reg}}^{\text{hinge}} = K\Bigl( [\max(0, \tau_o(k) - \hat{y}_o)]^2 + [\max(0, \hat{y}_t - \tau_t(k))]^2 \Bigr)
\end{equation}
Thresholds $\tau_o(k), \tau_t(k)$ tighten dynamically from $2\times$ slack down to nominal thresholds ($\tau_o=10\,\text{mm}, \tau_t=20\,\text{mm}$) over $T=1000$ steps. Step projections enforce trust-region bounds: position drift $\|\mathbf{p}-\mathbf{p}_0\|_\infty \le 20\,\text{mm}$, tangential growth $\le 30\%$, and contact-axis shrinkage $\le 90\%$. Periodic analytical snapping occurs every $K_{\text{snap}}=50$ iterations.

\begin{algorithm}[ht]
\caption{Gradient-Based Pairwise Geometry Repair}
\label{alg:pair_repair}
\begin{algorithmic}[1]
\Require Parent $(\mathbf{s}_P, \mathbf{p}_P)$ frozen; Child $(\mathbf{s}_0, \mathbf{p}_0)$; GAT model $f_\theta$; Learning rate $\eta = 3\times 10^{-3}$; Budget $T=1000$.
\State $(\mathbf{s}, \mathbf{p}) \gets (\mathbf{s}_0, \mathbf{p}_0)$; Infer contact axis $a^\parallel$
\State Probe $f_\theta(\mathbf{s}_P, \mathbf{p}_P, \mathbf{s}_0, \mathbf{p}_0)$; if $\hat{y}_o \ge \tau_o$, freeze position gradients $\nabla_{\mathbf{p}} \gets 0$
\For{$k = 1, \ldots, T$}
  \State $(\hat{y}_o, \hat{y}_t, \hat{\ell}_{\text{feas}}) \gets f_\theta(\mathbf{s}_P, \mathbf{p}_P, \mathbf{s}, \mathbf{p})$
  \State Compute $\mathcal{L}_{\text{repair}}$ and update $(\mathbf{s}, \mathbf{p})$ via Adam
  \State Project $(\mathbf{s}, \mathbf{p})$ onto trust-region box
  \If{$k \bmod K_{\text{snap}} == 0$}
    \State Snap $\mathbf{p}$ to parent contact face; reset Adam moments
  \EndIf
  \If{$k \bmod K_{\text{stop}} == 0$ \textbf{and} $\hat{y}_o > \tau_o$ \textbf{and} $\hat{y}_t < \tau_t$ \textbf{and} $P(\text{feas}) > 0.9$}
    \State \textbf{break}
  \EndIf
\EndFor
\State Final analytical snap onto contact face; \Return $(\mathbf{s}, \mathbf{p})$
\end{algorithmic}
\end{algorithm}

\section{MLP + KD-Tree Baseline Formulation}
\label{app:mlp_baseline}
The MLP baseline uses a unified 12-dimensional feature vector (3D relative position, 6D block dimensions, and 3D absolute Z coordinate) mapped through 3 hidden layers of width 128 (ReLU, dropout 0.35). The optimization loss incorporates a KD-tree distance penalty $\mathcal{P}_{\text{kd}}$ relative to valid training configurations:
\begin{equation}
\begin{split}
    \mathcal{L}_{\text{mlp}}(\mathbf{x}) ={} & -\log P(\text{Assembly\_Good} \mid \mathbf{x}) \\
    & + \lambda_{\text{kd}} \max\bigl(0, d(\mathbf{x}) - \tau\bigr)^2
\end{split}
\end{equation}
where $\lambda_{\text{kd}} = 0.2$ and distance threshold $\tau = 1.5$. Optimization uses 10 random-restart Adam trajectories with soft parameter clipping and multi-parent joint updates.

\end{document}